\documentclass[letterpaper]{article} 
\usepackage{aaai25}  
\usepackage{times}  
\usepackage{helvet}  
\usepackage{courier}  
\usepackage[hyphens]{url}  
\usepackage{graphicx} 
\usepackage{natbib}  
\usepackage{caption} 
\usepackage{algorithm}
\usepackage{algorithmic}
    
\usepackage{tikz}
    \usetikzlibrary{positioning,shadows}
    \tikzset{
        l/.style={rectangle,draw},
        n/.style={l,rounded corners},
        st/.style={rectangle,draw,fill=green!40!white}
    }
    \tikzstyle{level 1}=[level distance=10mm,sibling distance=10mm]

\usepackage{booktabs}
\usepackage{amsfonts}

\usepackage{newfloat}
\usepackage{listings}
\DeclareCaptionStyle{ruled}{labelfont=normalfont,labelsep=colon,strut=off} 
\floatstyle{ruled}
\newfloat{listing}{tb}{lst}{}
\floatname{listing}{Listing}
\usepackage{xspace}
\usepackage{algorithm}
\usepackage{algorithmic}
    
\usepackage{tikz}
    \usetikzlibrary{positioning,shadows}
    \tikzset{
        l/.style={rectangle,draw},
        n/.style={l,rounded corners},
        st/.style={rectangle,draw,fill=green!40!white}
    }
    \tikzstyle{level 1}=[level distance=10mm,sibling distance=10mm]
\usepackage{todonotes} 

\colorlet{mycyan}{cyan!20!white}
\colorlet{mypink}{pink!40!white}

\newtheorem{example}{Example}
\newtheorem{problem}{Problem}

\newcommand{\CSt}[1]{{\color{black!50!black}#1}}

\newcommand{\ALt}[1]{{\color{black}#1}}

\newcommand{\medict}[1]{{\color{black}#1}}

\newcommand{\R}{\ensuremath{\mathbb{R}}\xspace}
\newcommand{\N}{\ensuremath{\mathbb{N}}\xspace}
\newcommand{\B}{\ensuremath{\mathbb{B}}\xspace}
\newcommand{\states}{\ensuremath{\mathcal{S}}\xspace}
\newcommand{\predicates}{\ensuremath{\mathcal{P}}\xspace}
\newcommand{\A}{\ensuremath{A}\xspace}
\newcommand{\E}{\ensuremath{E}\xspace}

\newcommand{\spec}{\ensuremath{P}\xspace}
\newcommand{\controllers}[1][\A]{\ensuremath{\mathcal{C}_{#1}}\xspace}
\newcommand{\traces}{\ensuremath{\mathcal{T}}\xspace}
\newcommand{\BB}{\ensuremath{B}\xspace}
\newcommand{\minv}{\ensuremath{v_0}\xspace}
\newcommand{\inc}{\ensuremath{v_+}\xspace}

\newcommand{\answerPartial}[1][]{\textcolor{orange}{\bf [Partial]}}

\title{In Search of Trees:\\Decision-Tree Policy Synthesis for Black-Box Systems via Search}

\author{
    Emir Demirovi{\'c}\textsuperscript{\rm 1},
    Christian Schilling\textsuperscript{\rm 2},
    Anna Lukina\textsuperscript{\rm 1}
}

\affiliations{
    \textsuperscript{\rm 1} Delft University of Technology, The Netherlands\\
    \textsuperscript{\rm 2} Aalborg University, Denmark\\
    e.demirovic@tudelft.nl,
    christianms@cs.aau.dk,
    a.lukina@tudelft.nl
}

\begin{document}

\maketitle

\begin{abstract}
Decision trees, owing to their interpretability, are attractive as control policies for (dynamical) systems. Unfortunately, constructing, or synthesising, such policies is a challenging task. 
Previous approaches do so by imitating a neural-network policy, approximating a tabular policy obtained via formal synthesis, employing reinforcement learning, or modelling the problem as a mixed-integer linear program. However, these works may require access to a hard-to-obtain accurate policy or a formal model of the environment (within reach of formal synthesis), and may not provide guarantees on the quality or size of the final tree policy. 
In contrast, we present an approach to synthesise \emph{optimal} decision-tree policies given a \emph{deterministic black-box environment and specification}, a discretisation of the tree predicates, and an initial set of states, where optimality is defined with respect to the number of steps to achieve the goal. Our approach is a specialised search algorithm which systematically explores the (exponentially large) space of decision trees under the given discretisation. The key component is a novel trace-based pruning mechanism that significantly reduces the search space. Our approach represents a conceptually novel way of synthesising small decision-tree policies with \emph{optimality} guarantees even for \emph{black-box environments with black-box specifications}.
\end{abstract}

\begin{links}
    \link{Code}{https://doi.org/10.5281/zenodo.14601859}
\end{links}

\section{Introduction}

\begin{figure}[t]
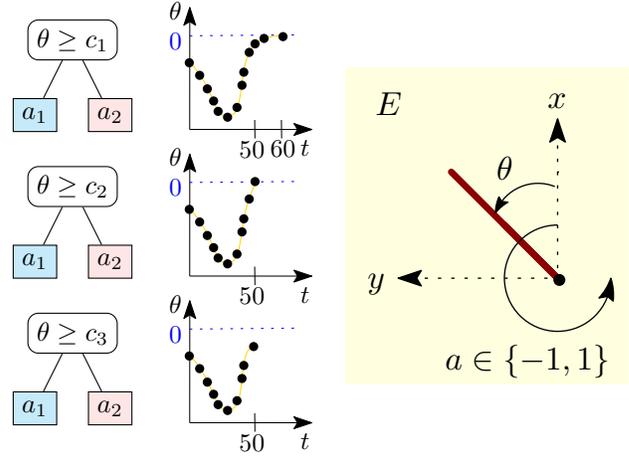

    \centering
    \include{pendulum_example}
    \caption{\ALt{Problem illustration. For the pendulum environment $E$ (right), each tree (left) with predicates $\theta\geq c_i,\:i=1,2,3,$ coupled with the black-boxed $E$, produces a trace for $\theta$ plotted along the time axis (middle) together with its evaluation with respect to the black-box specification (reaching $\theta = 0$). The middle trace is deemed best since it reaches the goal faster ($t=50<60$). The bottom trace is inferior.}
    }
    \label{fig:pendulum-example}
\end{figure}

Designing controllers for complex systems with the guarantee of specified behaviour remains an important challenge. Classical control synthesis can provide such guarantees given a (precise) model of the system~\cite{belta2019formal}. This requirement may in some cases be infeasible, which gave rise to black-box and approximate approaches, e.g., based on machine learning. As systems grow larger, interpretability is an increasingly desired specification for \ALt{machine-learned} 
policies \CSt{to achieve} \ALt{alignment with human specifications}~\cite{rudin2019stop}. With the success of decision trees as interpretable machine-learning models, policies represented as decision trees have gained considerable traction~\cite{DuLH20,glanois2024survey}.

There are diverse approaches to synthesising, or learning, decision-tree policies. Stratego~\cite{david2015uppaal} employs reinforcement learning dedicated to decision trees.  Modifying the reinforcement learning process to produce decision trees has also been proposed \cite{topin2021iterative}. An alternative is to apply imitation learning to distil a neural-network policy into a decision tree~\cite{BastaniPS18}. After using \CSt{formal} synthesis to construct a policy in tabular form, decision trees may be induced \ALt{via specialised algorithms akin to algorithms used for solving standard classification problems}~\cite{ashok2021dtcontrol}.

While previous approaches have their strengths, none of the discussed methods
provide guarantees in terms of decision-tree policy performance or size and/or require an existing expert policy or effective reinforcement learning algorithm. \ALt{Policy synthesis may be posed as a mixed-integer linear programming problem~\cite{vos2023optimal}, which can provide guarantees. However, this approach}
assumes that a model of the environment is given. When the above requirements are not met, decision-tree policy cannot be constructed using existing methods.

In contrast, we consider a unique setting: derive 1)~\emph{small} decision-tree policies when 2)~the model and specification of the environment is a \emph{deterministic black-box} whilst 3)~providing \emph{optimality guarantees} of performance, under the discretisation of tree predicates and initial states. \ALt{Every} work we are aware of will violate at least one of these three points.

Our work is for \emph{deterministic black-box systems},
%
\ALt{which pose a challenging controllability problem, as exact guarantees imply searching in an exponentially large space.}

\ALt{Our approach is based on search. Briefly, our algorithm systematically enumerates all possible decision trees that may be constructed using a given set of predicates, and then selects the tree that optimises the specification evaluated by the black-box environment for each tree, e.g., minimises (maximises) the time to reach (maintain) a target state.}

\begin{example}
As an illustrative example, consider the \emph{pendulum} environment in Fig.~\ref{fig:pendulum-example}. The pendulum is attached at one end to a fixed point, and available control actions are to apply force to push the free end left ($a_1=-1$) or right ($a_2=1$)~\cite{anderson1989learning}. Our aim is to construct a small decision-tree policy that swings the pendulum \CSt{to an upward angle} ($\theta=0$) from a given initial state, and does so as quickly as possible. The environment is available as a black box, i.e., the dynamics are hidden, \ALt{but given \CSt{an initial state and a policy,} we may compute the trajectory and obtain its evaluation with respect to the black-box \CSt{specification}}.

\CSt{We start using the first tree (with predicate~$[\theta\geq c_1]$ and leaf nodes corresponding to actions)}
as a policy for the black-box environment~$\E$\CSt{, and} obtain a trace that reaches the goal angle within $60$ time steps. Next, the predicate is modified to~$[\theta\geq c_2]$ (where $c_2>c_1$), and the new tree produces a trace that reaches the goal within $50$ time steps, which is considered better. For the next tree with predicate~$[\theta\geq c_3]$ (where $c_3>c_2$), the (partially) produced trace is considered inferior:
it surely does not reach the goal faster than the best tree (50 steps).
\end{example}

A key component is our novel \emph{trace-based pruning} mechanism that discards a large portion of the search space by runtime analysis. It exploits the decision-tree structure by considering the execution of the tree policy: \emph{even though the environment is black-box}, examining the trace allows us to understand \emph{how} the decision tree is used, and discard trees that are guaranteed to not lead to a better trace. This allows us to reduce the search space \emph{without} sacrificing optimality even though our model and specification are given as a black box. In the previous example, depending on the concrete trace, our trace-based pruning might be able to determine that it is possible to discard the third tree from consideration without missing a better tree only by observing the trace produced by the first and second tree. In practice this can lead to order-of-magnitude reductions in runtime.

We implement\CSt{ed} our approach and evaluate it on classical control benchmarks. The experiments demonstrate significant reductions obtained with our trace-based pruning, and illustrate that small and optimal decision trees may be constructed within reasonable time. We further analyse the scalability of our algorithm in terms of the number of predicates (granularity) and the size of the tree, both of which have an exponential influence on the runtime. The experiments show the runtime is within practical use.

To summarise, we consider a unique setting and provide a conceptually novel approach to construct optimal small decision-tree policies with respect to black-box systems. While not all environments are controllable by small trees, when the environment \emph{does} admit a small tree policy, our approach provides an effective way to compute an optimal tree only requiring black-box access to the system.


We organise the paper as follows. In the next section, we discuss related work and highlight our unique setting. \CSt{We outline preliminaries in Section~\ref{sec:preliminaries},} define the problem in Section~\ref{sec:problemdef}, present our approach in Section~\ref{sec:contributions}, experimentally evaluate our approach in Section~\ref{sec:experiments}, provide further discussion in Section~\ref{sec:discussion}, and conclude in Section~\ref{sec:conclusion}.




\section{Related Work}
\label{sec:relatedwork}
Our work covers a unique setting: constructing decision trees given a deterministic black-box system whilst providing optimal performance guarantees. There are no directly \CSt{applicable} works that we are aware of \CSt{in this setting}. To illustrate the challenges, we discuss works for synthesising decision-tree policies, albeit not fitting into our setting.


\emph{Reinforcement learning}. \ALt{A tree policy can be obtained via reinforcement learning, either by using dedicated tree algorithms~\cite{david2015uppaal} or by modifying reinforcement learning to output tree policies~\cite{topin2021iterative}.
Alternative approaches allow linear functions on the leaves~\cite{gupta2015policy}, consider multiple predicates, branches and actions at a time, or fix the structure using expert knowledge~\cite{likmeta2020combining} and then employ policy gradient updates~\cite{silva2020optimization,paleja2022learning}. These approaches perform exceptionally well, when an existing reinforcement-learning approach is available that is effective for the given system~\cite{topin2021iterative}, and/or \CSt{the} model~\cite{gupta2015policy} of the environment is known. A tree policy may be derived by imitating an expert policy, e.g., a neural network \cite{BastaniPS18}.
In contrast, we require neither model nor expert policy and provide optimality guarantees. In case of sparse rewards, reinforcement learning might struggle, while our framework by design has no such problem.}


\emph{Learning from tabular data}. When the policy is given in tabular form, dedicated tree-learning algorithms for control policies can be employed~\cite{ashok2020dtcontrol,ashok2021dtcontrol}, which extend classical tree-learning algorithms~\cite{BreimanFOS84,Quinlan96}. Recent advancements in optimal tree induction could potentially also be employed~\cite{demirovic2022murtree,van2024necessary}. However, obtaining the tabular policy requires an explicit model, which is not required in our setting.

\emph{Optimal policy synthesis}. The problem of constructing a tree policy may be posed as a mixed-integer linear program~\cite{vos2023optimal}, after which off-the-shelf solvers may be used to obtain optimal policies. However, not all environments may be feasible to model with such an approach (e.g., differential equations or trigonometric functions), and in our setting we consider black-box environments, which are not amendable to linear programming.

\emph{Verification}. There has been recent work to provide guarantees for decision-tree policies~\cite{SchillingLDL23}, possibly for infinitely many traces (we consider finite\CSt{ly many} traces). However, the tree policy and the model \CSt{must be given explicitly, which makes this work orthogonal to ours.}

We do note that we consider \emph{deterministic} systems with discrete actions. Some of the methods above are also applicable to stochastic environments and continuous actions, which we consider as future work.

To summarise, while there has been considerable work on decision-tree policies, synthesising such policies when the environment is black-box whilst also providing optimality guarantees is an open challenge.

%





\section{Preliminaries}
\label{sec:preliminaries}

A \emph{state}~$S = (s_1, s_2, \dots, s_d)^\top \in \states$ is a $d$-dimensional real-valued vector from a bounded state space~$\states \subseteq \R^d$, where each \emph{state dimension}~$s_i$ belongs to an interval~$s_i \in [\ell_i, u_i]$.
An action~$a \in \A$ comes from a finite set~$\A \subseteq \mathbb{Z}$ of integer-valued actions. An \emph{environment} is a function~$\E \colon \states \times \A \to \states$ that takes as input a state~$S$ and an action~$a$ and computes a \ALt{trajectory until asked to output the next \emph{observable}} state~$S' = \E(S, a)$. In our setting, the environment is treated as a black box, i.e., we are agnostic to its dynamics.

A \emph{policy} 
is a function~$\pi \colon \states \to \A$ that chooses an action based on an input state. We write~$\controllers$ for the set of all policies over action set~$\A$. 
We are concerned with the special case of \emph{a decision-tree policy}, which is given in the form of a binary tree where each inner node is called a \emph{predicate node} and each leaf node is called an \emph{action node}. Each predicate node is associated with a function~$\states \to \B$, where~$\B$ is the Boolean domain. Each action node is associated with one of the available actions~$a \in \A$. Given a state~$S$ and a node of the tree, a decision-tree policy~$\pi$ computes the action~$a = \pi(S)$ using the following recursive procedure, starting at the root node. If the current node is an action node, the associated action is returned. Otherwise, the current node is a predicate node. If the state~$S$ makes the predicate evaluate to true (false), the procedure continues with the left (right) child node.

Given an environment~$\E$, a decision-tree policy~$\pi$, an initial state~$S_0$, and a bound~$k \in \N$, the \emph{trace} with $k$ steps is the sequence of observed states $\tau = S_0, S_1, \dots, S_k$, obtained by applying the sequence of actions given by $\pi$: $S_i = \E(S_{i-1}, \pi(S_{i-1}))$, for $i=1,\ldots,k$. We write~$\traces$ for the set of all traces.

\section{Optimal Decision-Tree Policies}
\label{sec:problemdef}


We focus on the case of a single initial state~$S_0$, and generalise the problem to multiple initial states in Section~\ref{section:extensions}.

\subsection{Discretised Decision-Tree Predicates}
\label{subsection:predicates}

The space of all decision trees is infinitely large. By discretising the tree predicates, we obtain a finite (but exponentially large) space. We restrict our attention to (axis-aligned) predicates of the form $[s_i \geq \minv + m \cdot \inc]$, where~$s_i$ is the \emph{i-th} state dimension, $\minv, \inc \in \R$ are real-valued constants, and~$m \in \N$ is a positive integer. Since our state space is bounded, we obtain a finite number of nonequivalent predicates. We write~$\predicates$ for the set of all (tree) \emph{predicates}.

For instance, given a \CSt{state space with}~$s_i \in [0, 3]$ and~$\minv = \inc = 1$, we consider the predicates $[s_i \geq 1]$, $[s_i \geq 2]$, and $[s_i \geq 3]$. Note that predicate $[s_i \geq 0]$ is excluded since it is a tautology by~$s_i$'s domain, i.e., \CSt{always evaluates to} \emph{true}.

Considering a finer discretisation yields more predicates, which increases the space of considered decision trees, but allows potentially finding better trees. The algorithm's runtime is sensitive to the size of the search space, and hence a practical balance is needed. Experimentally we show that our approach can handle reasonably small increments.


\subsection{Specification}

\ALt{Given a deterministic environment~$\E$, decision-tree policy $\pi$, and initial state~$S_0$, we consider a specification to determine whether $\pi$ satisfies the specification. We assume that the specification is given in terms of traces, again in the form of a black-box \emph{specification function}~$\spec \colon \traces \to \B$.
In order to effectively determine whether $\pi$ satisfies the specification, we restrict the class of specifications we consider. A \emph{bounded-time prefix-closed specification} with a bound~$k \in \N$}
has the property that
every trace~$\tau$ of length~$k$ either satisfies or violates the specification (``bounded-time''), and whenever a trace~$\tau$ yields either of these verdicts, then any longer trace with the prefix~$\tau$ yields the same verdict (``prefix-closed''). As a consequence, 
we are guaranteed to obtain a \CSt{Boolean} verdict from a trace of length at most~$k$.
We call the (unique) trace of length~$k$ the \emph{witness trace}.

This class of specifications includes common reach-avoid problems where a goal needs to be reached while undesired states need to be avoided.
For instance, for the pendulum environment, the specification is to reach the vertical position within a step bound~$k$.
A trace satisfies the specification if and only if a prefix of length less than~$k$ satisfies the specification function. Conversely, any trace not reaching the goal withing~$k$ steps violates the specification.

\subsection{Optimality}

So far, we were only interested in finding \emph{any} policy that satisfies a given specification. In general, there may exist multiple solutions. We are interested in identifying \CSt{an} \emph{optimal} policy. For that, we assume a \emph{fitness function}, which is a partial order~${\succeq} \colon \traces \times \traces$ to compare two traces. A trace that satisfies the specification always precedes a trace that violates the specification. The fitness function induces another partial order~${\succeq} \colon \controllers \times \controllers$ to compare two policies. We say that $\pi_1$ is strictly better than $\pi_2$, written~$\pi_1 \succ \pi_2$, if one of the following conditions holds:
1)~The witness~$\tau_1$ has strictly better fitness than the witness~$\tau_2$ (i.e., $\tau_1\succ\tau_2$).
2)~Both witnesses have the same fitness, and~$\pi_1$ is a strictly smaller tree.

We wrap the black-box environment~$\E$ and the black-box specification function~$\spec$ into a black-box system~$\BB \colon \controllers \times \states \to \B \times \traces$. This system takes as input a policy~$\pi$ and an initial state~$S_0$ and \CSt{outputs both the} (Boolean) verdict and the trace~$\tau$. We say that a policy~$\pi$ satisfies the specification for initial state~$S_0$ if~$\BB$ yields a positive verdict. We note that~$\BB$ can be implemented from~$\E$ and~$\spec$ by simply generating the witness~$\tau$
and querying the specification function.

\begin{problem}\label{prob:main}
    Given a black-box system~$\BB$ \CSt{over} a set of actions~$\A$, an initial state~$S_0$, a limit on the depth and number of predicate nodes, a discrete set of predicates, and a fitness function~$\succeq$, find a decision-tree policy~$\pi \in\controllers$ within the defined size that satisfies the specification optimally with respect to the fitness function~${\succeq}$ and the witness trace~$\tau$ produced by the black-box system~$\BB$.
\end{problem}

For instance, for environment $\E$ in Fig.~\ref{fig:pendulum-example}, the black-box specification (reaching the vertical position) is satisfied for two out of three decision trees. However, we are looking for those trees that satisfy the specification within the smallest number of time steps (in this example, $50$).

\section{Synthesis of Optimal Decision-Tree Policies}
\label{sec:contributions}

For computing an \emph{optimal} decision-tree policy that solves Problem~\ref{prob:main}, a naive procedure is to enumerate all possible decision trees and evaluate them. By fixing an upper limit of the number of nodes and considering a discrete set of predicates, this procedure terminates. However, the number of trees is exponential, rendering this procedure infeasible. Our contribution is an efficient instantiation of this procedure.

\subsection{Searching In the Space of Decision Trees}




Our algorithm to enumerate decision trees is based on backtracking search. We represent the search space using \emph{backtracking variables}~$b_i$, where each variable is associated with a node in the tree. The possible values that can be assigned to a variable depend on the node type which the variable is associated with: predicate nodes may be assigned a predicate from the set of discretised predicates, whereas action nodes may be assigned an action from the set of available actions.

Backtracking variables are considered in a predefined order, i.e., variable $b_i$ goes before variable $b_{i+1}$. As is standard in backtracking, all combinations for variable $b_{i+1}$ are exhausted before taking the next value for variable $b_i$.

Assigning all backtracking variables results in a decision-tree policy. Consequently, enumerating all possible assignments to the backtracking variables corresponds to all possible policies in our available space. When enumerating a policy, it is used in combination with the black-box environment to compute the witness trace and evaluate the quality of the policy. Finally, the best policy is returned as the result.

For a tree with a fixed shape \CSt{and} $n$ predicate nodes, $|\A|$ number of actions, \CSt{and} $|\predicates|$ number of discretised predicates, the size of the total search space is $\mathcal{O}(|\predicates|^{n} \cdot |\A|^{n+1})$. Our contribution reduces this large search space in practice.


\subsection{Intuition Behind Trace-based Pruning}

\label{subsection:trace-pruning}

The idea of pruning is to limit the exploration by avoiding to explicitly enumerate trees that are guaranteed to be suboptimal, \CSt{i.e.}, do not satisfy the desired property with a higher fitness. We define sufficient conditions for pruning.

\begin{figure}[t]
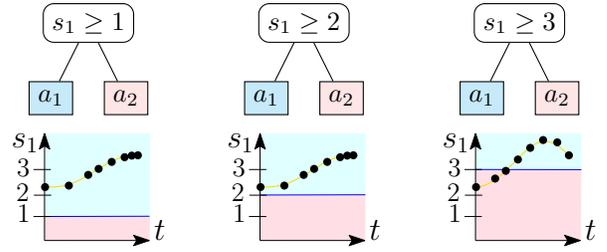

\centering
    \input{example_tree_1}
    \quad
    \input{example_tree_2}
    \quad
    \input{example_tree_3}
    \caption{Three decision-tree policies of fixed structure (top) and corresponding black-box traces (bottom). The decision boundary (blue) splits the state space into two regions (light blue and red). Predicate $s_1\geq 2$ can be skipped as the trace lies above the decision boundary.}
    \label{fig:exampleTrace}
\end{figure}

To provide the intuition behind our approach, consider an environment with only one state dimension $s_1$, and the process of enumerating all trees with exactly one predicate node, three possible predicates $[s_1 \geq 1]$, $[s_1 \geq 2]$, $[s_1 \geq 3]$, and having the left and right child nodes \emph{fixed} to actions $a_1$ and $a_2$, respectively (cf.\ Figure~\ref{fig:exampleTrace}). To find the best tree of the given description, we generally need to consider all three trees, with each tree differing only in the root node.

Assume the algorithm starts with predicate~$[s_1 \geq 1]$ and computes the first trace depicted in Figure~\ref{fig:exampleTrace}. We could consider \CSt{the} two remaining trees with predicates~$[s_1 \geq 2]$ and~$[s_1 \geq 3]$. However, from the \CSt{first} trace we see that the \CSt{second} tree with predicate~$[s_1 \geq 2]$ would result in the exact same trace. Indeed, there is no state in the trace where the policy would decide differently regardless of whether the predicate is~$[s_1 \geq 1]$ or~$[s_1 \geq 2]$, so the trace would not change. As a result, we do not have to evaluate the tree with predicate~$[s_1 \geq 2]$, and can directly go to the last tree.

\subsection{Trace-Based Pruning}

\label{subsection:trace-pruning2}

The intuition discussed above can be generalised to prune a \CSt{potentially} exponential number of trees that do not result in a different trace, which significantly speeds up the search process. In the following, we discuss incorporating a general version within a backtracking algorithm with more than one predicate node. We will consider predicates in increasing order, i.e., $[s_j \geq v_1]$ before~$[s_j \geq v_2]$ if~$v_1 < v_2$.

Given a backtracking variable~$b_i$ associated with a predicate node, the algorithm needs to select the next predicate to assign to the variable. A naive approach would be to simply select the next larger threshold, e.g., from the example in Section~\ref{subsection:trace-pruning}, assign predicate $[s_1 \geq 2]$ after considering $[s_1 \geq 1]$. However, we can leverage information about previous traces to avoid explicitly considering predicates that are guaranteed to not result in a trace that has not been previously observed. 

After assigning a predicate $[s_j \geq v]$ to a backtracking variable $b_i$, the idea is to track the values of state dimension~$s_j$ that have been observed during environment runs such that the predicate evaluated to \emph{true}. In particular, we are interested in the smallest such value, which we refer to as the distance value $d_i \in \R$. Note that a tree policy is only run after having all backtracking variables assigned.

The key idea is that, when selecting the next predicate for backtracking variable~$b_i$, its threshold value ($v$ in the example above) should \emph{exceed} the distance value $d_i$. Otherwise, the trace would be identical.

To reiterate, for each backtracking variable $b_i$ associated with a predicate node with current predicate $[s_j \geq v]$, we store a value $d_i \in \R$, which tracks the minimum value of state dimension~$s_j$ for which the predicate was evaluated to \emph{true} amongst all traces that were considered since the predicate $[s_j \geq v]$ has been assigned. Note that selecting a new threshold value for the predicate that is \emph{smaller or equal} than $d_i$ is guaranteed to result in traces already observed. Note that it is not necessary to track the values where the predicate evaluates to \emph{false}, since the predicates are explored in increasing order of the thresholds and as such the future predicates would also evaluate to false on those values.

Initially, the distance value $d_i$ is set to \emph{undefined} each time a new predicate $[s_j \geq v]$ is assigned as part of \CSt{the} search. The first time the node observes a state where its predicate is satisfied in a trace, the distance value $d_i$ is set to the corresponding value of state dimension $s_j$. Each subsequent time the predicate is satisfied, $d_i$ is updated to the smallest value for which the predicate still evaluates to \emph{true}.

After considering predicate $[s_j \geq v]$ for node $i$, our algorithm does not consider the next predicate, but instead uses the predicate $[s_j \geq v']$ where~$v'$ is the smallest available value such that~$v' > d_i$.
If the distance value~$d_i$ is undefined, \emph{all} predicates may be discarded for that backtracking variable for the currently considered state dimension $s_j$.

For the example in Section~\ref{subsection:trace-pruning},
the distance value $d_1$ is initially undefined, and upon completing the first trace, it is set to $d_1 = 2.3$. When selecting the next predicate, $[s_1 \geq 2]$ is discarded since its threshold $2$ does not exceed distance value $d_1 = 2.3$; so the next selected predicate is $[s_1 \geq 3]$.

The above idea is applied to every backtracking variable associated with a predicate.

Our pruning strategy is computationally inexpensive: it amounts to tracking a single value for each backtracking variable, and updating this value as the tree is queried during trace computation. The algorithm retains completeness, as it is guaranteed to not discard optimal trees. Our trace-based pruning is the key component in the practical efficiency.

\subsubsection{Additional Techniques:}

\textbf{Trees explored in increasing size.} The algorithm partitions the search space in terms of tree shapes, which are ordered by size. For example, after considering trees with exactly one predicate node, the algorithm considers trees which have a root node with one left predicate child, then trees which have a root node with one right predicate child, then complete trees with three predicate nodes, and so on until the size budget is reached.


\textbf{Early stopping due to the objective.} During the search, the best tree found so far is tracked. When evaluating a new candidate tree, its evaluation is preemptively stopped when it is determined that the trace cannot be extended to a trace that is better than the one obtained from the best tree so far.

For example, consider a setting where the policy should reach a goal state as quickly as possible. If the best policy so far reaches the goal state in~$k$ trace steps, and the partial trace associated with the current candidate policy has not reached the goal state in~$k-1$ steps, we may safely discard the candidate policy from further consideration, since it cannot result in a better trace. Note that, since we explore trees in increasing size, this results in the algorithm computing the smallest tree with the optimal performance across the considered maximum tree-size budget.


Early stopping has two advantages: 1)~it saves computational time, and 2)~it results in traces of shorter length, which allows for more aggressive trace-based pruning \medict{owing to fewer distance updates being made}.


\textbf{Symmetries.} Trees that have identical left and right subtrees are discarded from consideration. In these cases, the root node of such a tree is redundant and its predicate has no influence \CSt{on} the trace. A smaller tree, consisting of the subtree, would result in the same trace, and since the algorithm explores trees in increasing size, it is safe to discard the larger symmetric tree without further consideration. The main use of this technique is to discard trees that contain predicate nodes with identical left and right action nodes, and otherwise plays a minor role.

\subsubsection{Summary.}

\begin{algorithm}[t]
	\caption{Construction of an optimal decision-tree policy}
	\label{alg:reach}
	\begin{algorithmic}[1]
        \REQUIRE An initial state $S_0$, black-box system $B$, predicates~$\predicates$, fitness~$\succeq$, tree-size bound
        \ENSURE Policy $\pi$ that produces the trace with the best fitness function
        \STATE bvar $\leftarrow$ NextBacktrackingVariable() \label{line:start}
        \STATE val $\leftarrow$ NextValue(bvar) \hfill \COMMENT{use trace-based pruning if predicate node}
        \STATE Assign(bvar, val)
        \IF{assignment failed, all values exhausted}
            \STATE Backtrack()
            \IF{not possible to backtrack}
                \RETURN{$\pi_{global}$} \label{line:return}
            \ENDIF
        \ENDIF
        \IF{all backtracking variables assigned}
            \STATE $\pi_{local} \leftarrow$ ConstructTreePolicy(bvars)
            \STATE $\tau_{local} \leftarrow B(\pi, S_0)$ 
            \IF{$\tau_{local} \succ \tau_{global}$}
                \STATE $\tau_{global} \leftarrow \tau_{local}$
                \STATE $\pi_{global} \leftarrow \pi_{local}$
            \ENDIF
            \FOR{bvar in bvars}
                \STATE n $\leftarrow \pi_{local}$.nodes[bvar.index]
                \STATE bvar.distance $\gets \min($n.distance, bvar.distance$)$
            \ENDFOR
        \ENDIF
        \STATE \textbf{Goto} Line~\ref{line:start}.
	\end{algorithmic}
\end{algorithm}

Algorithm~\ref{alg:reach} provides a high-level view on using backtracking variables. If available, the next unassigned backtracking variable is selected, or the last assigned variable otherwise. The next value is selected either as the next action for variables representing action nodes, and otherwise trace-based pruning is used to determine the threshold. Once a predicate has been exhausted on one state dimension, predicates for the next state dimension are selected.

Once all backtracking variables are assigned, the algorithm constructs a decision-tree policy, and uses the black-box system to produce the trace~$\tau$. If~$\tau$ is better than the globally best trace (initially \CSt{\emph{null}}), that trace is updated to~$\tau$. The distance values of the nodes of the policy are used to update the distance values of the backtracking variables. After all policies have been (implicitly) considered, the algorithm returns the best policy (Line~\ref{line:return}).


\subsection{Extensions}
\label{section:extensions}

\textbf{Multiple initial states.} The previous discussion was based on constructing a tree policy from a single initial state. However, we may be interested in finding a single tree policy that works well across multiple initial states. The algorithm remains similar, with impact on two components: 1)~the objective function, and 2)~trace-based pruning.

When evaluating a tree with respect to multiple initial states, we generalise the fitness function. For example, if the goal is to minimise the trace length, then the generalisation aims to minimise the maximum trace length. This influences early stopping: the initial states are evaluated with respect to the tree policy one at a time, and as soon as a trace is encountered that is considered violating, the evaluation stops, i.e., the remaining initial states are not considered further. 

The above idea interacts with trace-based pruning. In case the tree evaluation is stopped early, meaning the tree is deemed not better than the best tree found so far, only the last trace is used to update the distance values. The intuition is that, if we wish to find a better tree, it must \CSt{lead to} a trace different from the last trace, and we can ignore the distance-value updates of the other traces.

As a result, due to the interaction with pruning, finding an optimal tree with respect to multiple initial states may lead to sub-linear runtime with respect to the number of states.



\textbf{Maximisation.} Rather than satisfying the desired property in the least number of steps, we may be interested in maximising the number of steps. For example, the goal may be to balance a pole for as long as possible. The algorithm stays largely the same, with the only analogous changes needed in the evaluation of the tree. For maximisation it is important to specify an upper bound on the trace length; otherwise, the algorithm may potentially run infinitely long.

\section{Experimental Study}
\label{sec:experiments}

We aim to illustrate the effectiveness of our approach with a proof-of-concept implementation. We show that our trace-based pruning approach is a key factor in making the approach feasible. Furthermore, we consider scalability from two perspectives: the granularity of the predicates, and the number of predicate nodes in the tree. While both are expected to have an exponential impact on the runtime, we observe that the runtime is still within practical use.


We consider three classical control problems: CartPole, MountainCar, and pendulum. The environment behaviour is defined as in Gymnasium\footnote{https://github.com/Farama-Foundation/Gymnasium}, with the adjustment that we maximise (CartPole - at most 10k steps) or minimise (MountainCar and pendulum) the trace length rather than use their reward functions.
The control actions are limited to two choices, e.g., apply maximum force in one or the other direction. Our implementation only \CSt{queries} the environment in a black-box fashion. We generated the initial states randomly within a specified range; see the Appendix for details about the parameters, environment description, and a sample of trees produced by our approach.



To reiterate, as discussed in Section \ref{sec:relatedwork}, we work with a unique setting where we 1) synthesise \emph{decision-tree} policies, 2) only require \emph{black-box} \CSt{access to} the \emph{deterministic} system, and 3) provide \emph{guarantees} on performance under the tree definition, e.g., the policy that minimises the time taken, or prove that no such policy exists. Every work we are aware of violates at least one of these points. Consequently, while direct comparisons with other works may be done by relaxing the requirements of our setting, this brings considerable caveats that result in comparisons that we argue are not meaningful with respect to our contribution. For these reasons, we focus on demonstrating the feasibility of our approach and its scalability.

We implemented our approach `Broccoli' in Rust 1.77.0. The experiments were run on consumer-grade hardware (Xeon(R) W-10855M @ 2.8 GHz). The code is public.


\subsection{Experiment \#1: Trace-Based Pruning}

\begin{table}[ht!]
    \caption{Runtime and variance with and without trace-based pruning. Pruning substantially reduces runtime. Multiple initial states demonstrate only a sub-linear runtime increase.}
    \label{tab:results-runtime}
    \centering
    \begin{tabular}{%
    @{}l%
    @{}c%
    r@{}l@{}r@{}l%
    @{}r}
    \toprule
&\multicolumn{1}{c}{}&\multicolumn{4}{c}{Runtime (seconds)}&\multicolumn{1}{c}{No.\ of trees}\\
Environment%
&\multicolumn{1}{c}{$|S_0|$}%
&\multicolumn{2}{c}{Not Prune} & \multicolumn{2}{c}{Prune}%
&\multicolumn{1}{c}{(Not) Prune} 
\\
\cmidrule(r){1-1}
         \cmidrule(r){2-2}
         \cmidrule(r){3-6}
         \cmidrule(r){7-7}
         CartPole&$1$&$3.13$&\scriptsize{$\pm1.79$}&$0.22$&\scriptsize{$\pm0.20$}&{\scriptsize$(2.2m)$} $49k$\\
MountainCar&$1$&$4.72$&\scriptsize{$\pm0.49$}&$0.21$&\scriptsize{$\pm0.04$}&{\scriptsize$(518k)$} $20k$\\
Pendulum&$1$&$13.75$&\scriptsize{$\pm11.07$}&$3.19$&\scriptsize{$\pm3.03$}&{\scriptsize$(845k)$} $126k$\\
\cmidrule(r){1-1}
         \cmidrule(r){2-2}
         \cmidrule(r){3-6}
         \cmidrule(r){7-7}
CartPole&$100$&$7.46$&\scriptsize{$\pm1.28$}&$1.28$&\scriptsize{$\pm0.35$}&{\scriptsize$(2.2m)$} $88k$\\
MountainCar&$100$&$6.55$&\scriptsize{$\pm0.68$}&$0.40$&\scriptsize{$\pm0.08$}&{\scriptsize$(518k)$} $24k$\\
Pendulum&$100$&$131.4$&\scriptsize{$\pm9.41$}&$23.93$&\scriptsize{$\pm2.46$}&{\scriptsize$(845k)$} $205k$\\
\bottomrule
\end{tabular}
\end{table}

\begin{table*}[t]
    \caption{Runtime increases as we increase the number of predicates per state dimension (X), with the exception of CartPole, where finer discretisation sometimes finds a perfect tree faster. Variance remains similar as in Table \ref{tab:results-runtime} proportional to runtime. }
    \label{table:granular}
    \centering
    \begin{tabular}{%
    @{}l%
    r@{}l@{\hspace{5pt}}r@{}l@{\hspace{5pt}}r@{}l@{\hspace{5pt}}r@{}l
    r@{\hspace{5pt}}r@{\hspace{5pt}}r@{\hspace{5pt}}r}%
    \toprule
&\multicolumn{8}{c}{Runtime (seconds) and variance}&\multicolumn{4}{c}{No.\ of trees }\\
Environment%
&\multicolumn{2}{c}{X=5} &\multicolumn{2}{c}{X=10} &\multicolumn{2}{c}{X=15} &\multicolumn{2}{c}{X=20}%
&\multicolumn{1}{c}{X=5} &\multicolumn{1}{c}{X=10} &\multicolumn{1}{c}{X=15} &\multicolumn{1}{c}{X=20}%
\\
\cmidrule(r){1-1}
         \cmidrule(r){2-9}
         \cmidrule(r){10-13}
         CartPole&$102$&{\scriptsize$\pm128$}&$1420$&{\scriptsize$\pm2696$}&$10.2$&{\scriptsize$\pm16.33$}&$287$&{\scriptsize$\pm668$}&$25m$&$334m$&$1m$&$46m$\\
MountainCar&$6.26$&{\scriptsize$\pm8.55$}&$14.0$&{\scriptsize$\pm3.86$}&$95.49$&{\scriptsize$\pm29.2$}&$446$&{\scriptsize$\pm97$}&$30k$&$1.3m$&$9.5m$&$45m$\\
Pendulum&$6.58$&{\scriptsize$\pm7.53$}&$30.75$&{\scriptsize$\pm61.32$}&$113$&{\scriptsize$\pm151$}&$366$&{\scriptsize$\pm501$}&$18k$&$1.2m$&$15m$&$51m$\\
\bottomrule
\end{tabular}    
\end{table*}

\begin{table*}[t]
    \caption{Runtime and performance (not depicted) increase as we increase the number of nodes (N).}
    \label{tab:numnodes}
    \centering
    \begin{tabular}{%
    @{}l%
    r@{}l@{\hspace{5pt}}r@{}l@{\hspace{5pt}}r@{}l@{\hspace{5pt}}r@{}l
    r@{\hspace{5pt}}r@{\hspace{5pt}}r@{\hspace{5pt}}r}%
    \toprule
&\multicolumn{8}{c}{Runtime (seconds) and variance}&\multicolumn{4}{c}{No. of trees}\\
Environment%
&\multicolumn{2}{c}{N=3} &\multicolumn{2}{c}{N=4} &\multicolumn{2}{c}{N=5} &\multicolumn{2}{c}{N=6}%
&\multicolumn{1}{c}{N=3} &\multicolumn{1}{c}{N=4} &\multicolumn{1}{c}{N=5} &\multicolumn{1}{c}{N=6}%
\\
\cmidrule(r){1-1}
         \cmidrule(r){2-9}
         \cmidrule(r){10-13}
         CartPole&$0.1$&{\scriptsize$\pm0.01$}&$1.16$&{\scriptsize$\pm0.11$}&$11.85$&{\scriptsize$\pm7.26$}&$39.92$&{\scriptsize$\pm38.45$}&$27k$&$294k$&$2.9m$&$9.4m$\\
MountainCar&$0.12$&{\scriptsize$\pm0.02$}&$0.74$&{\scriptsize$\pm0.16$}&$5.37$&{\scriptsize$\pm1.16$}&$14.1$&{\scriptsize$\pm3.39$}&$9k$&$62k$&$484k$&$1.2m$\\
Pendulum&$0.3$&{\scriptsize$\pm0.21$}&$0.56$&{\scriptsize$\pm0.23$}&$2.17$&{\scriptsize$\pm0.41$}&$5.36$&{\scriptsize$\pm0.83$}&$7k$&$44k$&$309k$&$826k$\\
\bottomrule
\label{table:numnodes}
\end{tabular}    
\end{table*}




We run experiments with and without our trace-based pruning (Table~\ref{tab:results-runtime}), both for one and~$100$ initial states, for trees of depth two. Initial states are generated randomly, and \CSt{the results are} averaged over ten runs.

Our trace-based pruning technique is clearly effective in pruning the search space. For CartPole and MountainCar, there is a one- to two-orders-of-magnitude difference, whereas for pendulum, it leads to a 4x reduction.

The runtime is roughly proportional to the number of trees explicitly considered, as expected, and is consistent based on the standard deviation across different initial states. Interestingly, we observe only a sub-linear increase in runtime with more initial states, and the total number of trees considered is roughly similar regardless of the number of initial states. 

\subsection{Experiment \#2: Granularity of Predicates}


The granularity of predicates impacts the runtime: the finer the discretisation, the larger the search space. Each state dimension in the environment has a predefined range of values that it may take. We divide this interval into 5, 10, 15, and 20 values, and use these values as the predicate thresholds.

Table~\ref{table:granular} summarises the results for trees of depth three across randomly generated initial states (averaged over ten runs). The general trend is that increasing the number of predicates indeed increases the runtime.

However, in the case of CartPole, we observe an opposite effect: finer predicates \emph{decrease} the runtime. This is because the finer discretisation resulted in the algorithm finding a tree faster that balances the pole for $10{,}000$ steps, which is the maximum number of steps so the search can terminate.

The results indicate that the discretisation needs to be chosen carefully to balance runtime and quality of the final tree.

\subsection{Experiment \#3: Number of Predicate Nodes}



The number of predicate nodes influences the search space size. We study the runtime for trees of depth three and varying the maximum number of nodes from three to six. 

The results are summarised in Table~\ref{table:numnodes}. Note that the search space of trees with at most $k$ predicate nodes \emph{includes} trees with less than $k$ predicate nodes, meaning the search space is strictly larger as we increase the number of nodes. Consequently, there is a sharp increase in the runtime, which is expected due to the exponential factor.

\section{Further Discussion and Limitations}
\label{sec:discussion}

\ALt{
Our approach is effective at computing small and optimal decision-tree policies given a discretisation of the predicates and a set of initial states. There is an exponential runtime dependency with respect to the size of the tree and the discretisation of the state space. It may be infeasible with our approach to construct large tree policies or deal with high-dimensional environments. It is also possible that not all environments may be controlled by small decision trees.

However, when it \emph{is} applicable, we believe small trees are valuable for interpretability reasons, and our approach provides the means to easily obtain such trees.
The exponential runtime factor in our approach is inherent to every approach that aims to provide guarantees.

Our approach is exceptionally flexible as it only requires black-box access to the system. This entails that the black box may be arbitrarily complex, as long as it can still be practically computed. Furthermore our algorithm \CSt{provides performance} guarantees, despite working with black boxes.

The optimality is important since it guarantees that we obtain the best performing tree given our definition, which may be relevant for some applications. It also allows us to conclude \CSt{in cases when} no such tree exists, and in general understand the limits of decision trees as control policies.


Given that our approach is a conceptually novel way to synthesise decision-tree policies in a unique setting, \CSt{it} opens many avenues for future work.

Parallelisation is promising as the search space can be naturally partitioned, and further \emph{heuristic} pruning may lead to a principled trade-off between runtime and guarantees. Extending the approach to stochastic environments is another interesting direction. In our work, continuous actions ought to be discretised in a preferably smaller number of actions. Synthesising optimal trees for continuous actions remains an open challenge for decision-tree policies in general.

\section{Conclusion}
\label{sec:conclusion}

We presented a novel search-based method for computing an optimal decision-tree policy given a black-box deterministic system, a set of initial states, and a discretisation of the tree predicates. To the best of \CSt{our} knowledge, our approach is the first to consider such a setting. The key component \CSt{is} our trace-based pruning technique, which discards large portions of the search space at runtime. We illustrated the practicality of the approach on classical control benchmarks. When the environment is controllable by a small tree, our approach provides a way to obtain a small and optimal tree despite only requiring black-box access to the system.
}



\section{Acknowledgements}
This research was funded by the NWO Veni grant Explainable Monitoring (222.119) and by the Independent Research Fund Denmark under reference number 10.46540/3120-00041B. This work was done in part while Anna Lukina was visiting the Simons Institute for the Theory of Computing.

\bibliography{aaai25}


\clearpage
\appendix

\section{Appendix / supplemental material}

\subsection{\!Proof of Correctness for Trace-Based Pruning}

We introduced trace-based pruning in Sections \ref{subsection:trace-pruning} and \ref{subsection:trace-pruning2}. Here we provide a proof sketch of the idea.

\emph{Reminder.} For a given backtracking variable, predicates are explored in increasing order of the threshold. For example, after the predicate $[s_1 \geq v]$ has been assigned to a backtracking variable, after backtracking, any predicate that would be considered as the next predicate for the backtracking variable must have a greater threshold value, i.e., have the form $[s_1 \geq v']$ with $v' > v$.

In principle, we could explore arbitrary orderings of predicates for a given backtracking variable. For correctness, any ordering would do, as long as we explore all predicates. However, for our trace-based pruning to be effective, it makes book-keeping simpler if we order the predicates by increasing threshold value.

The order of backtracking variables is fixed in our algorithm and does not change during the execution. The variables are indexed by this order, i.e., all combinations for variable $b_{i+1}$ are exhausted before taking the next value for backtracking variable $b_i$. The distance $d_i$ for backtracking variable $b_i$ with corresponding predicate $[s_j \geq v]$ is the minimum value of the state dimension $s_j$ for which the predicate was evaluated to \emph{true} amongst all traces that were considered since the predicate $[s_j \geq v]$ has been assigned.

Our pruning mechanism is designed for the case when considering all possible trees of a fixed shape, i.e., the positions of the predicate and leaf nodes are fixed, and the goal is to enumerate all such trees. Recall that in our algorithm, we enumerate all tree shapes, and for each shape, we enumerate all corresponding trees where we make use of our trace-based pruning.

\emph{Claim.} Let us consider a single initial state. Let the backtracking variable $b_i$ have a predicate $[s_j \geq v]$ assigned with distance value $d_i$. When all predicates for all backtracking variables $b_k$ with $k > i$ have been exhausted, the goal is to assign the next predicate to backtracking variable $b_i$. The claim is that any predicate of the form $[s_j \geq v']$ with $v' < d_i$ will \emph{not} result in a tree policy that is better than the current best tree policy. In order words, the algorithm should assign a predicate $[s_j \geq v'']$ with $v'' = \min\{v : v > d_i\}$, or consider predicates with the next state dimension if no such value exists, or further backtrack if that is also not possible.

\emph{Proof sketch.} To see why trace-based pruning is correct, we give a proof by contradiction. Assume that there is a tree policy~$\pi^*$ that is better than the current best known policy~$\pi^+$, and that~$\pi^*$ has all nodes associated with backtracking variables $b_m$ with $m < i$ fixed to the same values as above, the node associated with the backtracking variable $b_i$ has a predicate $[s_j \geq v'']$ with $v'' < d_i$, and the remaining nodes are assigned any predicates. 

This means that the trace produced by~$\pi^*$ has at least one state such that, when querying~$\pi^*$ for the next action, the tree evaluates the predicate associated with the backtracking variable $b_i$ and $s_j < d_i$. However, this is not possible: since the search explored all possible traces obtainable with the allowed predicates, by definition of~$d_i$ there cannot be a trace for which~$\pi^*$ queries the node $b_i$ with $s_j < d_i$. This concludes the proof by contradiction.

\subsection{Environments}\label{app:models}

We considered three environments as defined in the Gymnasium\footnote{https://github.com/Farama-Foundation/Gymnasium}. The dynamics may be found in their git repository, however, our algorithm does not have access to the internal dynamics, and only observes the state-action outputs in a black-box fashion.
In the following, we describe the initial states and specifications. Note that specifications are also treated as black-box for the algorithm. 

\subsubsection{Reward function.}

Our goal is to maximise or minimise the trace length, i.e., the time it takes to reach the goal. Note that this is different from typical reinforcement learning, where the reward function generates a value at each state, and the objective is to minimise or maximise the cumulative reward. In this sense, that reward function is relatively smooth, whereas in our work, the reward is essentially binary: either the goal has been reached or not.

\subsubsection{CartPole.}

The system has four dimensions: cart position $x$, cart velocity $\dot{x}$, pole angle $\theta$, and pole angular velocity $\dot{\theta}$. We select the initial state by randomly assigning values in the range $[-0.05, 0.05]$ to each state dimension. These values follow the initial states given in the Gymnasium.

The specification is to maintain that the cart position stays within the range $[-2.4, 2.4]$ and the pole angle is within the range $[-\alpha, \alpha]$, where $\alpha = 24 \pi / 360$, by applying force to the left ($-1$) and right ($1$).

The goal is to maximise the trace size, up to a maximum of $10{,}000$ steps (other works usually consider up to $200$ steps).

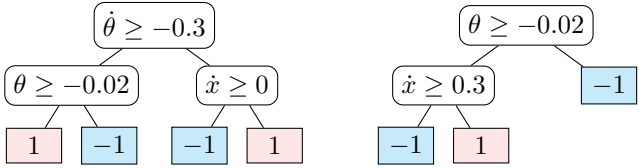
\begin{figure}[h!]
    \centering
    \begin{tikzpicture}[level 1/.style={level distance=8mm,sibling distance=22mm},level 2/.style={level distance=8mm,sibling distance=10mm}]
	\node[n] {$\dot{\theta} \geq -0.3$}
		child{ node[n] {$\theta \geq -0.02$}
			child{ node [l,fill=mypink,text width=5mm,text centered] {$1$}}
			child{ node [l,fill=mycyan,text width=5mm,text centered] {$-1$}}
		}
		child{ node[n] {$\dot{x} \geq 0$}
			child{ node [l,fill=mycyan,text width=5mm,text centered] {$-1$}}
			child{ node [l,fill=mypink,text width=5mm,text centered] {$1$}}
		};
\end{tikzpicture}
\hfill
\begin{tikzpicture}[level 1/.style={level distance=8mm,sibling distance=22mm},level 2/.style={level distance=8mm,sibling distance=10mm}]
	\node[n] {$\theta \geq -0.02$}
		child{ node[n] {$\dot{x} \geq 0.3$}
            child{ node [l,fill=mycyan,text width=5mm,text centered] {$-1$}}
			child{ node [l,fill=mypink,text width=5mm,text centered] {$1$}}
		}
        child{ node [l,fill=mycyan,text width=5mm,text centered] {$-1$}};
\end{tikzpicture}
    \caption{Visualisation of example decision trees for depth two, predicate increments $\{0.1,0.1,0.1,0.1\}$, and random with seed $0$ initial state $[0.013, -0.02, 0.047, 0.025]$ (left) and $100$ randomly sampled initial states (right) with seed $9$.}
    \label{fig:cartpole-tree}
\end{figure}
    
\subsubsection{MountainCar.}

The system has two dimensions: car position $x$ and car velocity $\dot{x}$. We select the initial state by randomly assigning the car position in the range $[-0.6, -0.4]$ and setting the velocity to zero. These values follow the initial states given in the Gymnasium.

The specification is to reach the top of a hill, which means that the car position is greater or equal to $0.5$, by applying force to the left ($-1$) and right ($1$).

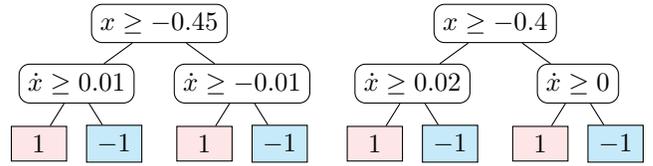
\begin{figure}[h!]
    \centering
    \begin{tikzpicture}[level 1/.style={level distance=8mm,sibling distance=22mm},level 2/.style={level distance=8mm,sibling distance=10mm}]
	\node[n] {$x \geq -0.45$}
		child{ node[n] {$\dot{x} \geq 0.01$}
			child{ node [l,fill=mypink,text width=5mm,text centered] {$1$}}
			child{ node [l,fill=mycyan,text width=5mm,text centered] {$-1$}}
		}
		child{ node[n] {$\dot{x} \geq -0.01$}
			child{ node [l,fill=mypink,text width=5mm,text centered] {$1$}}
            child{ node [l,fill=mycyan,text width=5mm,text centered] {$-1$}}
		};
\end{tikzpicture}
\hfill
\begin{tikzpicture}[level 1/.style={level distance=8mm,sibling distance=22mm},level 2/.style={level distance=8mm,sibling distance=10mm}]
	\node[n] {$x \geq -0.4$}
		child{ node[n] {$\dot{x} \geq 0.02$}
			child{ node [l,fill=mypink,text width=5mm,text centered] {$1$}}
			child{ node [l,fill=mycyan,text width=5mm,text centered] {$-1$}}
		}
		child{ node[n] {$\dot{x} \geq 0$}
			child{ node [l,fill=mypink,text width=5mm,text centered] {$1$}}
			child{ node [l,fill=mycyan,text width=5mm,text centered] {$-1$}}
		};
\end{tikzpicture}
    \caption{Visualisation of example decision trees for depth two, predicate increments $\{0.05,0.005\}$, and random with seed $0$ initial state $[-0.51,0]$ (left) and $100$ randomly sampled initial states (right) with seed $9$.}
    \label{fig:mountaincar-tree}
\end{figure}

\subsubsection{Pendulum.}

The system has two dimensions: the position of the free end of the pole $x$ and pole angular velocity $\theta$. Note that in the Gymnasium, the position of the tip of the pole is given as two dimensions in terms of $\cos$ and $\sin$ of the position - presumably this is done to make it easier for neural networks to learn; however, in our case, we chose to directly represent the position in Cartesian coordinates.

The initial state is given by randomly assigning values to the position of the tip of the pole and the angular velocity from the ranges $[-0.8, -0.5]$ and $[-0.2, 0.2]$, respectively. The values in Gymnasium are given within $[-1.0, 1.0]$, however, depending on the random values chosen for the experiments with multiple initial states, some configurations resulted in trees, whereas some did not, i.e., no decision tree within the given specification could reach the goal within $10{,}000$ steps. To make the experiments more consistent, we opted for a reasonable reduction of the initial states, and selected the above provided values for the initial states.

The specification is to reach a state where both state dimensions (in radians) are within the range $[-0.1, 0.1]$ by applying force to the left ($-1$) and right ($1$).

\begin{figure}[h!]
    \centering
    \begin{tikzpicture}[level 1/.style={level distance=8mm,sibling distance=22mm},level 2/.style={level distance=8mm,sibling distance=10mm}]
	\node[n] {$\theta \geq -5.2$}
		child{ node[n] {$x \geq -0.8$}
			child{ node [l,fill=mycyan,text width=5mm,text centered] {$-1$}}
			child{ node [l,fill=mypink,text width=5mm,text centered] {$1$}}
		}
		child{ node[n] {$\theta \geq -5.6$}
			child{ node [l,fill=mycyan,text width=5mm,text centered] {$-1$}}
			child{ node [l,fill=mypink,text width=5mm,text centered] {$1$}}
		};
\end{tikzpicture}
\hfill
\begin{tikzpicture}[level 1/.style={level distance=8mm,sibling distance=22mm},level 2/.style={level distance=8mm,sibling distance=10mm}]
	\node[n] {$\theta \geq 0.2$}
		child{ node[n] {$x \geq -0.2$}
            child{ node [l,fill=mycyan,text width=5mm,text centered] {$-1$}}
			child{ node [l,fill=mypink,text width=5mm,text centered] {$1$}}
		}
        child{ node[n] {$\theta \geq -1$}
            child{ node [l,fill=mycyan,text width=5mm,text centered] {$-1$}}
            child{ node [l,fill=mypink,text width=5mm,text centered] {$1$}}
        };
\end{tikzpicture}
    \caption{Visualisation of example decision trees for depth two, predicate increments $\{0.2,0.2\}$, and random with seed $0$ initial state $[-0.665, -0.024]$ (left) and $100$ randomly sampled initial states (right) with seed $9$.}
    \label{fig:pendulum-tree}
\end{figure}
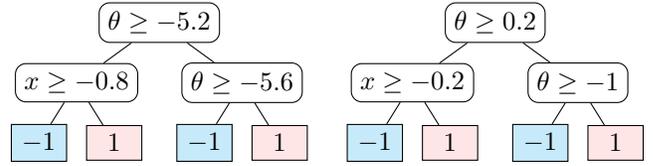



\end{document}